\documentclass[10pt, a4paper]{article}

\usepackage[final]{lrec2026}

\usepackage{booktabs}      
\usepackage{multirow}      
\usepackage{makecell}
\usepackage{graphicx}      
\usepackage[table]{xcolor} 
\definecolor{darkgreen}{rgb}{0.0, 0.5, 0.0}
\definecolor{darkblue}{rgb}{0.0, 0.0, 0.75}
\definecolor{darkred}{rgb}{0.75, 0.0, 0.0}
\usepackage{pgfmath}       
\usepackage{siunitx}       
\usepackage{amsmath}       
\usepackage{todonotes}
\usepackage{txfonts}
\usepackage{algorithm, algpseudocode}
\makeatletter
\newcommand{\LongState}[1]{%
  \State \parbox[t]{\dimexpr\linewidth-\ALG@thistlm\relax}{#1\strut}%
}
\makeatother
\newlength{\lrCellWidth}
\newlength{\lrCellHeight}
\definecolor{darkgreen}{rgb}{0.0, 0.0, 0.5}
\definecolor{lightgreen}{rgb}{0.0, 0.0, 1.0}
\definecolor{darkred}{rgb}{0.5, 0.0, 0.0}
\definecolor{lightred}{rgb}{1.0, 0.0, 0.0}
\definecolor{yellow}{HTML}{F28522}

\newcommand{\up}[1]{\textcolor{darkblue}{\scriptsize$\uparrow$#1\%}}
\newcommand{\down}[1]{\textcolor{darkred}{\scriptsize$\downarrow$#1\%}}
\newcommand{\celldata}[4]{%
  \begingroup
  \sisetup{detect-weight, round-mode=places, round-precision=1, table-format=+-1.3}%
  \pgfmathparse{#4 < 0.05 ? 1 : 0}\let\isbold\pgfmathresult
  \pgfmathparse{max(-50, min(50, #3))}\let\clampedpct\pgfmathresult
  \pgfmathparse{(\clampedpct + 50) / 100}\let\ratio\pgfmathresult
  \pgfmathparse{\ratio < 0.5 ? 2*\ratio : 1}\let\redval\pgfmathresult
  \pgfmathparse{\ratio < 0.5 ? 2*\ratio : 2*(1-\ratio)}\let\greenval\pgfmathresult
  \pgfmathparse{\ratio > 0.5 ? 2*(1-\ratio) : 1}\let\blueval\pgfmathresult
  \pgfmathparse{abs(#3) > 30 ? 1 : 0}\let\usewhite\pgfmathresult
  \colorbox[rgb]{\redval,\greenval,\blueval}{%
    \makebox[\lrCellWidth][c]{%
      \ifnum\usewhite=1\color{white}\else\color{black}\fi
      \ifnum\isbold=1 {\bfseries \num{#1}}\else\num{#1}\fi
      {\scriptsize\,$\pm$\,\num{#2}}%
    }%
  }%
  \endgroup
}

\usepackage{fancyhdr}

\usepackage[utf8]{inputenc}
\usepackage{makecell, supertabular}
\usepackage{booktabs}
\usepackage{multirow}

\usepackage{xcolor}
\definecolor{darkgreen}{rgb}{0.0, 0.5, 0.0}
\definecolor{darkblue}{rgb}{0.0, 0.0, 0.75}
\definecolor{darkred}{rgb}{0.75, 0.0, 0.0}

\usepackage{microtype}
\usepackage{supertabular}

\usepackage{fancyhdr}

\title{Simple Additions, Substantial Gains:\\Expanding Scripts, Languages, and Lineage Coverage in \textsc{URIEL+}}

\name{%
\begin{tabular}{@{}c@{}}
Mason Shipton\textsuperscript{$\vardiamondsuit$}\quad
York Hay Ng\textsuperscript{$\varheartsuit$}\quad
Aditya Khan\textsuperscript{$\varheartsuit$}\quad
Phuong Hanh Hoang\textsuperscript{$\varheartsuit$}\\
Xiang Lu\textsuperscript{$\spadesuit$}\quad
A.~Seza Doğruöz\textsuperscript{$\clubsuit$}\quad
En\mbox{-}Shiun Annie Lee\textsuperscript{$\vardiamondsuit\,\varheartsuit$}
\end{tabular}}

\address{%
\textsuperscript{$\vardiamondsuit$}Ontario Tech University\quad
\textsuperscript{$\varheartsuit$}University of Toronto\\
\textsuperscript{$\spadesuit$}University of Michigan\quad
\textsuperscript{$\clubsuit$}LT3, IDLab, Universiteit Gent\\
\texttt{masonshipton25@gmail.com}, \texttt{as.dogruoz@ugent.be},\\
\texttt{annie.lee@ontariotechu.ca}
}

\abstract{
The \textsc{URIEL+} linguistic knowledge base supports multilingual research by encoding languages through geographic, genetic, and typological vectors. However, data sparsity (e.g. missing feature types, incomplete language entries, and limited genealogical coverage) remains prevalent. This limits the usefulness of \textsc{URIEL+} in cross-lingual transfer, particularly for supporting low-resource languages.
To address this sparsity, we extend \textsc{URIEL+} by introducing script vectors to represent writing system properties for $7,488$ languages, integrating Glottolog to add $18,710$ additional languages, and expanding lineage imputation for $26,449$ languages by propagating typological and script features across genealogies. These improvements reduce feature sparsity by $14\%$ for script vectors, increase language coverage by up to $19,015$ languages ($1,007\%$), and boost imputation quality metrics by up to $35\%$. Our benchmark on cross-lingual transfer tasks (oriented around low-resource languages) shows occasionally divergent performance compared to \textsc{URIEL+}, with performance gains up to $6\%$ in certain setups. 
 \\ \newline  }

\begin{document}

\maketitleabstract
\thispagestyle{fancy}
\section{Introduction}

Linguistic knowledge bases enable us to understand the relationships between languages, in particular conducing the rise of cross-lingual transfer learning \cite{bjerva-etal-2020-sigtyp}. One such knowledge base is \textsc{URIEL}  \citelanguageresource{littell-etal-2017-uriel}, which represents languages with three vector types: geographic (i.e. based on $299$ coordinate points), genetic (i.e. family membership), and typological (i.e. syntactic, phonological, and inventory features). Through the lang2vec tool, \textsc{URIEL} provides pre-computed language distances. Downstream applications of these distances include cross-lingual transfer \cite{lin-etal-2019-choosing}, dependency parsing \cite{ustun-etal-2020-udapter}, and performance prediction (\citealp{khiu-etal-2024-predicting}, \citealp{anugraha-etal-2025-proxylm}).

\textsc{URIEL+} \citelanguageresource{khan-etal-2025-uriel} partially addresses data sparsity and transparency issues in \textsc{URIEL} \citelanguageresource{littell-etal-2017-uriel} as identified by \citet{toossi-etal-2024-reproducibility}, by expanding data coverage to over $8,000$ languages. Furthermore, \textsc{URIEL+} prevents distance computation between language pairs that lack shared features. Although this is restrictive, \textsc{URIEL+} allows users to circumvent this by advanced imputation methods to fill in missing values (e.g. by way of the SoftImpute algorithm, due to \citet{mazumder2010spectral}). \emph{Lineage imputation}, a method of partial imputation, optionally populates features for dialects of six languages (Spanish, French, English, German, Malay, and Arabic), given that phylogenetic relationships of languages are strong predictors of language similarity \cite{dunn2011nature}. 

\begin{figure*}
    \centering
    \includegraphics[width=1.0\linewidth]{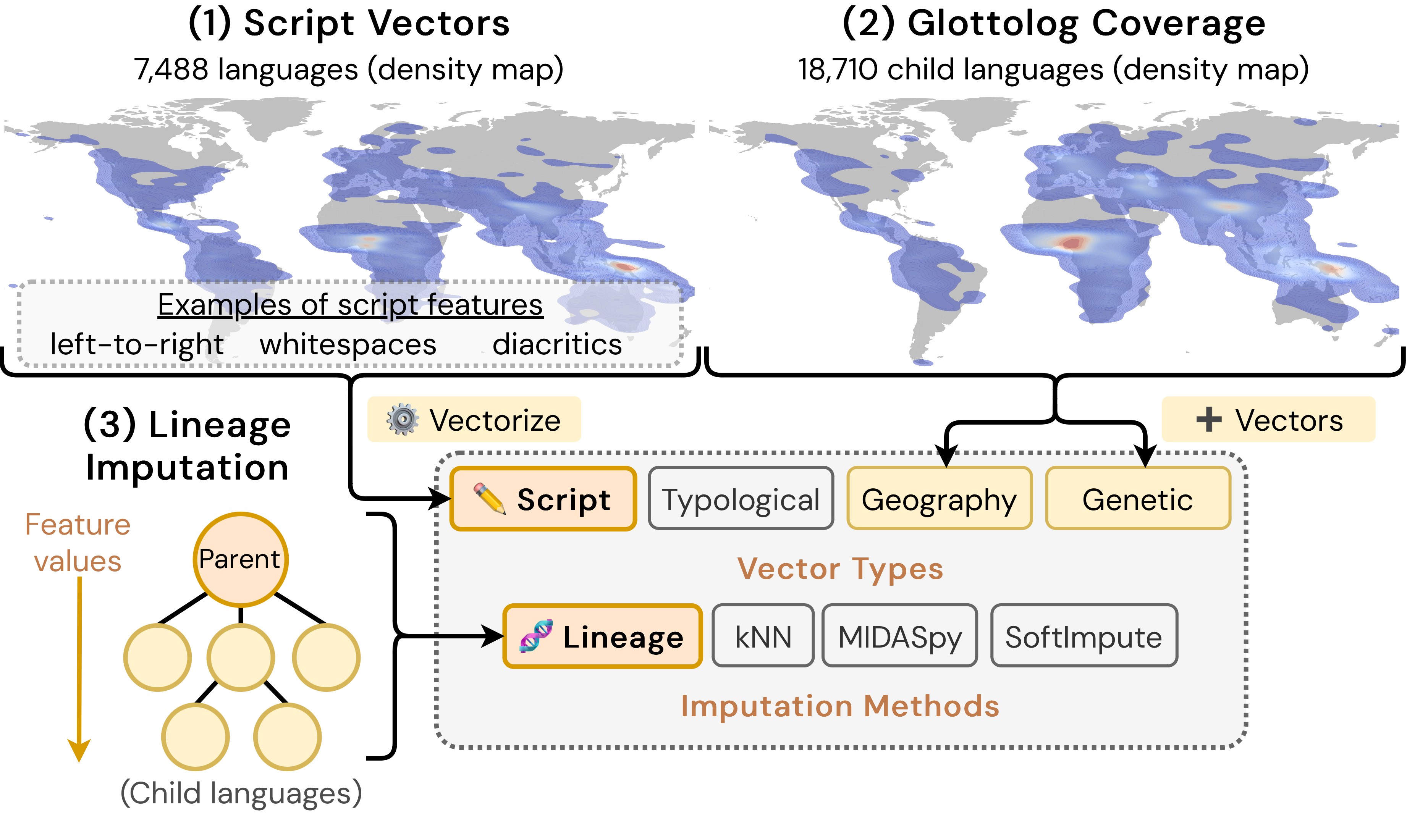}
    \caption{Overview of our additions to mitigating data sparsity in \textsc{URIEL+}. (1) Density map of $7,488$ languages with integrated script vectors. (2) Density map of the $18,710$ Glottolog child languages added to \textsc{URIEL+}. (3) Illustration of expanded lineage imputation, where missing values in child languages are filled from ancestors.}
    \label{fig:placeholder}
\end{figure*}

Nonetheless, three key limitations persist in \textsc{URIEL+}: (1) the absence of writing system representations, (2) incomplete language coverage (i.e. many languages catalogued in Glottolog \citelanguageresource{swj-glottocodes}, a widely used catalogue of the world's languages and language families, remain absent), and (3) the narrow scope of lineage imputation. Each limitation reflects a form of data sparsity (i.e. missing feature types, missing languages, and limited phylogenetic propagation). Our study addresses all limitations by extending \textsc{URIEL+} through (1) \textit{script vectors}, (2) \textit{Glottolog integration}, and (3) \textit{expanded lineage imputation} (see Figure~\ref{fig:placeholder} which summarises our three additions that collectively reduce sparsity and expand feature coverage in \textsc{URIEL+}). 

\begin{table*}[h]
\centering
\begin{tabular}{l c c c c c}
\hline
\textbf{Language Pair} & \textbf{Glottocodes} 
& \multicolumn{2}{c}{\textbf{Pre-Imputation}}
& \multicolumn{2}{c}{\textbf{Post-Imputation}} \\
\hline
 &  & \textbf{Script} & \textbf{Typology} & \textbf{Script} & \textbf{Typology} \\
\hline
Russian - Kazakh   & russ1263 - kaza1248 & 0.4788 & 0.6310 & 0.5000 & 0.6650 \\
Korean - Japanese  & kore1280 - nucl1643 & 0.2380 & 0.4706 & 0.2380 & 0.4827 \\
English - Turkish  & stan1293 - nucl1301 & 0.1864 & 0.6001 & 0.1864 & 0.6031 \\
Spanish - Finnish  & stan1288 - finn1318 & 0.0000 & 0.4949 & 0.0000 & 0.5054 \\
\hline
\end{tabular}
\caption{Languages similar in script (i.e. script distance closer to 0) but divergent in typology (i.e. featural distance closer to 1), shown before and after union aggregation of databases, and SoftImpute imputation.}
\label{tab:script_typo_imputation}
\end{table*}

Table \ref{tab:script_typo_imputation} shows language pairs that are expected to exhibit similarity in script but divergence in linguistic typology. Each pair shares a common script (e.g. Cyrillic for Russian and Kazakh, Latin for English, Turkish, Spanish, and Finnish).
Korean and Japanese do not currently share the same script. However, both languages historically incorporated Chinese characters, which should result in low script distances. Before and after imputation, URIEL+ produces the anticipated pattern of low script distances coupled with higher typological distances, confirming that shared script does not necessarily imply typological similarity and that script as a feature provides unique insights into languages that typological currently does not.

\paragraph{Contribution 1: Script Vectors}    
While \textsc{URIEL+} captures a wide range of linguistic dimensions, it omits information about writing systems. Script properties are necessary for encoding structural and functional information relevant to natural language processing (NLP) tasks. Differences in script types (e.g. alphabetic or abjadic\footnote{A writing system that represents individual sounds as symbols.}) result in challenges in tokenisation and morphological processing. Moreover, shared scripts often reflect historical or cultural contact between genetically distant languages (e.g. Arabic script for Persian and Urdu). An encoding of these relationships is missing in \textsc{URIEL+}, but it is important for cross-lingual transfer \cite{zhuang-etal-2025-enhancing}. Therefore, we introduce \textit{script vectors} to encode structural and social properties of writing systems for $7,488$ languages. Through analyses in feature coverage and sparsity, we highlight how vectors expand \textsc{URIEL+}’s coverage in a novel direction, particularly benefitting low-resource languages where script vectors are among the few reliably documented vector types.

\paragraph{Contribution 2: Glottolog Integration} A second form of sparsity arises from limited language coverage in \textsc{URIEL+}. While \textsc{URIEL+} includes over $8,000$ languages, Glottolog \citelanguageresource{swj-glottocodes} catalogues $18,710$ additional languages with well-documented genealogical information. By incorporating these missing languages, \textsc{URIEL+} contains a complete encoding of language families in Glottolog. Beyond increasing coverage, this integration enables a phylogeny-based method for imputing missing values, discussed below. 

\paragraph{Contribution 3: Expanded Lineage Imputation}
Lineage imputation in \textsc{URIEL+} fills missing values in child languages by copying feature values from their immediate parent. As it stands, it supports only $6$ medium- to high-resource parent languages. This reflects the empirical observation that related languages tend to share typological features \cite{Dunn2005, JagerWahle2021}. We extend this approach to systematically propagate typological and script features from parent languages to all child languages across the knowledge base, forming an \textit{expanded lineage imputation} method that benefits $26,449$ languages in total following Glottolog integration. Furthermore, we demonstrate that this method yields consistent performance gains in imputation quality.

Finally, we benchmark the impact of these contributions in choosing languages for cross-lingual transfer across a variety of tasks, a common use case of linguistic knowledge bases. The performance gains from ablations of our contributions highlight their practicality for cross-lingual transfer, while particularly supporting low-resource languages.

\section{Literature Review}
Current research shows strong support for our three contributions to \textsc{URIEL+}. The existing literature in NLP demonstrates the importance of script data for modelling language relationships, while linguistic studies show that genetically related languages tend to share typological structures. Together, these findings support our proposed additions, which aim to reduce data sparsity and improve coverage in \textsc{URIEL+}.

\begin{table*}[!htp]
\centering
\footnotesize
\setlength{\tabcolsep}{6pt}
\renewcommand{\arraystretch}{1.15}
\begin{tabularx}{\textwidth}{@{}l
  >{\raggedright\arraybackslash}p{.36\textwidth}
  >{\raggedright\arraybackslash}X@{}}
\toprule
ScriptSource Feature & Possible Values & Binarised \textsc{URIEL+} Feature(s) \\
\midrule
Type & Alphabet; Abjad; Abugida; Featural; Logo-syllabary; Syllabary
     & \texttt{SC\_ALPHABET, SC\_ABJAD, SC\_ABUGIDA, SC\_FEATURAL, SC\_LOGO\_SYLLABARY, SC\_SYLLABARY} \\
Case & Yes; No
     & \texttt{SC\_CASE} \\
Ligatures & Required; Optional; None
          & \texttt{SC\_LIGATURES, SC\_REQUIRED\_LIGATURES} \\
\bottomrule
\end{tabularx}
\caption{Three examples of how binarisation of ScriptSource features was performed using one-hot encoding.}
\label{tab:script-binarization}
\end{table*}

\subsection{Script Data Usage in Literature}\label{subsec:script-data-nlp}
The script of a language is a first-order determinant of language similarity, and it is crucial in many NLP tasks. 

For \emph{morphological inflection}, \citet{murikinati-etal-2020-transliteration} show that transfer between related languages degrades when scripts differ. Transliterating the source into the target script (or both into a common script) improves accuracy, sometimes outweighing typological relatedness. For \emph{transfer learning}, \citet{amrhein-sennrich-2020-romanization} and \citet{tufa-etal-2024-unknown} report that romanising the script of a child language improves transfer  performance when parent and child use different scripts.
For \emph{multilingual pre-training}, LANGSAMP \cite{liu-etal-2025-langsamp} adds learnable script embeddings to the transformer output during masked language modelling (MLM), yielding better zero-shot transfer and more reliable source-language selection. \citet{xhelili-etal-2024-breaking} further show that transliteration-based post-training alignment improves \emph{cross-lingual alignment}.  Without such alignment, token representations from different scripts are nearly linearly separable. Finally, \citet{zhuang-etal-2025-enhancing} find that script differences systematically hinder transfer in low-resource languages.

Prior approaches rely on ad hoc, language-specific pre-processing (e.g. romanisation per \citet{amrhein-sennrich-2020-romanization}) which are unevenly available and biased toward high-resourced languages. These issues with bias can be addressed by adding script vectors to \textsc{URIEL+}, which reduces reliance on pre-processing and improves modelling of linguistic similarity, especially for low-resource languages.

\subsection{Glottolog as a Language Resource}
Despite \textsc{URIEL+}'s extensive coverage of over $8,000$ languages, many languages remain missing. For example, Bajjika \cite{toossi-etal-2024-reproducibility} is missing in \textsc{URIEL} and \textsc{URIEL+}, while Western Armenian and Sakizaya are missing in \textsc{URIEL}  
\footnote{While Sakizaya has been added to \textsc{URIEL+} via the Grambank linguistic source \citelanguageresource{skirgaard2023grambank}, Western Armenian remains absent from \textsc{URIEL+}}
\cite{novotny-etal-2021-one}.
The missing languages  highlight the need for a more comprehensive resource to ensure inclusive representation in linguistic knowledge bases.

Glottolog provides a solution to this sparsity by offering a comprehensive, authoritative catalogue of the world’s languages, dialects, and language families \citelanguageresource{swj-glottocodes}, encompassing approximately $18,000$ more languages than \textsc{URIEL+}. It uses a standardised system of language codes (Glottocodes) and assigns language codes to each language family and language, including dialects. \textsc{URIEL+} uses these codes to integrate data from linguistic sources such as BDPROTO (ancient languages; \citelanguageresource{marsico-etal-2018-bdproto}), Grambank \citelanguageresource{skirgaard2023grambank}, and eWAVE (English dialects; \citelanguageresource{ewave}). Moreover, Glottolog provides evidence-based genealogical classifications, facilitating integration across linguistic datasets. Recognised as the authoritative source on the world’s languages, Glottolog has been used in high-impact NLP and computational linguistics research \citep{liang2023holisticevaluationlanguagemodels,bommasani2022opportunitiesrisksfoundationmodels}, demonstrating its value for addressing gaps in \textsc{URIEL+} and supporting more inclusive multilingual studies.

\subsection{Phylogenetic Feature Propagation}
Phylogenetic relationships of languages are strong predictors of structural similarity. \citet{Dunn2005} demonstrate that typological features carry a phylogenetic signal in Oceanic and Papuan languages. \citet{dunn2011nature} consider evolutionary rates of typological and lexical features, showing that slower-changing features track lineage more closely. \citet{JagerWahle2021} proposed methods to estimate typological feature distributions while controlling for shared ancestry, and \citet{Hubler2022} examined structural features, identifying which exhibit strong phylogenetic signal. 

These existing studies motivate us to extend \textsc{URIEL+}'s lineage imputation to infer missing typological and script features for all languages based on their parent languages. This approach reduces sparsity and enhances coverage across \textsc{URIEL+}.

\section{Method}
This section describes how we extend \textsc{URIEL+} to reduce data sparsity and broaden feature coverage via three steps: (1) introducing script vectors, (2) integrating languages from Glottolog, and (3) expanding lineage imputation.

\subsection{Adding Script Vectors}
We introduce \textit{script vectors} as a new vector type in \textsc{URIEL+} to represent structural and social properties of writing systems. Features were sourced from ScriptSource \citelanguageresource{Holloway-2012}, a database of scripts, their typologies, and language usage. ScriptSource data is publicly viewable but it is not directly accessible. However, with assistance from the one of the ScriptSource administrators, we obtained dataframes describing: (a) writing system properties for all scripts, (b) languages in ScriptSource with their ISO 639-3 codes, and (c) scripts available for each language.

The writing system properties from dataframe (a) were binarised as shown in Table~\ref{tab:script-binarization}. For example, the ScriptSource feature \emph{Type}, which includes values such as \emph{Alphabet}, \emph{Abjad}, \emph{Abugida}\footnote{A writing system where consonants carry inherent vowels.}; was expanded such that each value became its own binary feature (e.g. \texttt{SC\_ALPHABET}, \texttt{SC\_ABJAD}, \texttt{SC\_ABUGIDA}), with a value of $1$ (if present) or $0$ (if absent). This binarisation process is the same as it was done in \textsc{URIEL} \citelanguageresource{littell-etal-2017-uriel} and \textsc{URIEL+} \citelanguageresource{khan-etal-2025-uriel} for typological vectors. 
For languages that can be written in multiple scripts (e.g. Kazakh uses Cyrillic, Arabic, and Latin scripts), the union of feature values was applied. This was implemented as a logical OR operation, where a feature was marked as present ($1$) if it existed in any of the language's associated scripts. As an example, Kazakh is marked as having both the \texttt{SC\_ABJAD} feature (from Arabic) and the \texttt{SC\_ALPHABET} feature (from Cyrillic and Latin).

These script vectors were integrated alongside \textsc{URIEL+}’s genetic, geographic, and typological vectors. Language-to-language distances based on script features were computed using normalised angular distance:
\[
D_{\text{ang}}(A,B) = \frac{1}{\pi}\arccos\left(\frac{A \cdot B}{\|A\| \|B\|}\right)
\]
where lower values indicate higher similarity (e.g. French and Spanish using the Latin script). 
This addition fills a representational gap, providing a high-coverage source of data for low-resource languages in particular.

\subsection{Full Integration of Glottolog}

We integrate all $18,710$ remaining Glottolog languages into \textsc{URIEL+}, expanding genealogical and geographic coverage. More specifically, we retrieved a list of all languages from Glottolog in their Glottocodes. Languages from Glottolog that were already in \textsc{URIEL+} (over $8,000$ of them) were excluded. Glottolog was turned into a linguistic source in \textsc{URIEL+} \citelanguageresource{khan-etal-2025-uriel}, similar to the World Atlas of Language Structures (WALS; \citelanguageresource{dryer-haspelmath-2013-wals}), Syntactic Structures of the World's Languages (SSWL; \citelanguageresource{koopman-2009-sswl}), and Grambank \citelanguageresource{skirgaard2023grambank}, where it can be integrated by choice to add the new languages. Unlike the other linguistic sources, Glottolog does not include new typological or script features, and it does not include typological data. Although these languages have no typological and script data, nearly all of them possess genetic classifications, allowing them to benefit directly from the lineage-based imputation to reduce sparsity.

Geographic coordinates and family affiliations for the Glottolog languages were obtained using the \texttt{lingtypology} package \cite{moroz2017lingtypology}, as \textsc{URIEL+} did for their new languages. Although Glottolog contributes no new vector types, its inclusion provides the genealogical backbone required for large-scale feature propagation of \textsc{URIEL+}.

\subsection{Lineage Imputation Methodology}
To further reduce sparsity, we extend \textsc{URIEL+}’s \emph{lineage imputation} to all languages, applying it to both typological and script vectors. This approach assumes that parent languages and their children share highly similar structural and orthographic properties, a pattern empirically supported in \textsc{URIEL+} ($89.8\%$ agreement in feature values between parent and children for typological data, $97.0\%$ for script data). These high agreement rates motivate a systematic propagation of known feature values to fill gaps across languages. However, this propagation does not resolve all missing values. Instead, it serves as a structured preprocessing step that can complement subsequent imputation methods such as SoftImpute \cite{mazumder2010spectral}, where otherwise the imputation of empty language vectors yields an uninformative result.

\begin{figure*}[ht]
  \includegraphics[width=1.0\textwidth]{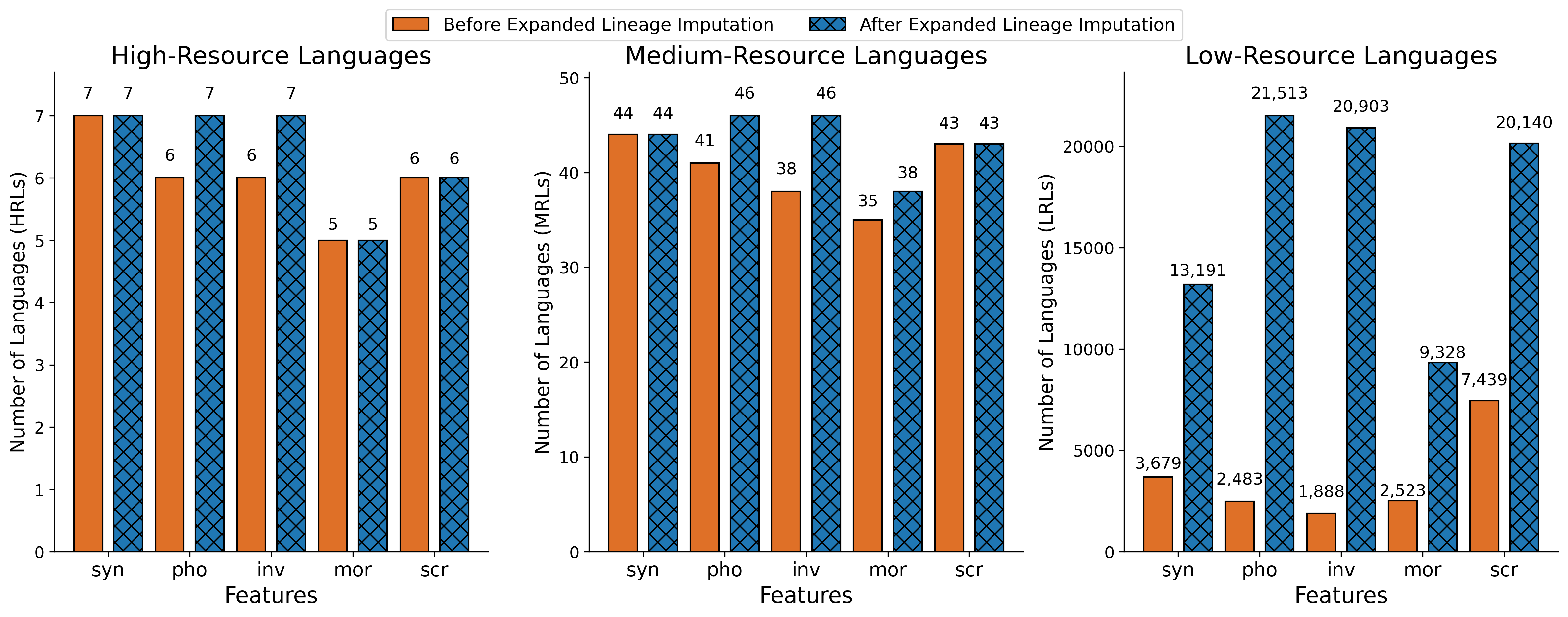}
   \caption{Number of languages with available syntactic (syn), phonological (pho), inventory (inv), morphological (mor), and script (scr) data in \textsc{URIEL+} before and after expanded lineage imputation. All linguistic sources in \textsc{URIEL+}, as well as Glottolog languages, are integrated. Shown for high-resource (HRL), medium-resource (MRL), and low-resource (LRL) languages \citep{joshi-etal-2020-state} from left to right.}
    \label{fig:feature_coverage_resource_level}
\end{figure*}

Algorithm \ref{algo:lineage} outlines the expanded lineage imputation algorithm. Parent-child mappings were derived from Glottolog \citelanguageresource{swj-glottocodes}, containing $13,372$ parent languages which support imputation for $26,449$ child languages after Glottolog integration. Relationships were modelled as a forest of language trees. Imputation was performed via breadth-first search (BFS). Feature values from parent nodes were propagated to immediate children (continuing until all descendants were filled). This ensures hierarchical, consistent propagation, improving \textsc{URIEL+} coverage and reducing sparsity at every phylogenetic level. The lineage imputation algorithm can be enabled or disabled if required, as described in the documentation.

\begin{algorithm}[H]
\caption{Expanded Lineage Imputation via Breadth-First Search (BFS)}
\label{algo:lineage}
\begin{algorithmic}[1]
\State Construct a mapping of Glottocodes from parent to child languages.
\State Define root languages as all parent languages who have no parent.
\ForAll{root language $r$}
    \LongState{Initialise queue $q \gets [r]$ and visited set $s \gets \{r\}$.}
    \While{$q$ not empty}
        \State Dequeue language $p$.
        \ForAll{child language $c$ of $p$}
            \If{$c \notin s$}
                \LongState{Let $f_p, f_c$ be binary feature vectors for $p,c$.}
                \LongState{For all $i$, set $f_c[i] \gets f_p[i]$ if $f_c[i] = \text{NaN}$ and $f_p[i] \neq \text{NaN}$.}
                \State Enqueue $c$ to $q$ and add $c$ to $s$.
            \EndIf
        \EndFor
    \EndWhile
\EndFor
\end{algorithmic}
\end{algorithm}

\section{Analysis of Knowledge Base Statistics}
This section examines how our additions affect the structure and completeness of \textsc{URIEL+}. We focus on (1) feature coverage (to measure the breadth of linguistic data across feature types and resource levels), (2) sparsity (to assess the proportion and distribution of missing values before and after expanded lineage imputation), and (3) correlations between script distances and other language distances (to evaluate the distinctiveness of script-based information). This analysis summarises how we make \textsc{URIEL+} less sparse, thereby augmenting the availability of data for downstream applications.

\subsection{Low-Resource Languages Have Higher Feature Coverage}

We measured feature coverage in \textsc{URIEL+} before and after expanded lineage imputation, counting the number of languages with at least one feature with data for syntactic, phonological, inventory, morphological, and script features. All five linguistic sources in \textsc{URIEL+} and Glottolog languages were integrated. Languages were categorised as high-, medium-, or low-resource (HRL, MRL, LRL) following \citet{joshi-etal-2020-state}.

Figure~\ref{fig:feature_coverage_resource_level} shows that script vectors substantially improve baseline coverage, providing data for $7,488$ languages before expanded lineage imputation (which is nearly double that of syntactic vectors ($3,730$ languages)). This difference is driven by gains in LRLs coverage. Syntactic vectors cover $7$ HRLs, $44$ MRLs, and $3,679$ LRLs. Meanwhile, script vectors cover $6$ HRLs (\textcolor{darkred}{$\downarrow 17\%$}), $43$ MRLs (\textcolor{darkred}{$\downarrow 2\%$}), and $7,439$ LRLs (\textcolor{darkblue}{$\uparrow 102\%$}). Although script vectors include marginally fewer HRLs and MRLs, they reach nearly twice as many LRLs, highlighting their capacity to alleviate sparsity where documentation is most limited.

Expanded lineage imputation amplifies these gains for LRLs across all vector types. Morphological vectors show the smallest absolute increase, adding $6,805$ languages (\textcolor{darkblue}{$\uparrow 270\%$}) due to sparse documentation. Syntactic vectors gain $9,512$ (\textcolor{darkblue}{$\uparrow 259\%$}) languages. Phonological, inventory, and script vectors exhibit the largest gains, adding $19,030$ (\textcolor{darkblue}{$\uparrow 766\%$}), $19,015$ (\textcolor{darkblue}{$\uparrow 1,007\%$}), and $12,701$ (\textcolor{darkblue}{$\uparrow 171\%$}) languages, respectively, consistent with their genealogical stability. All three vector types now cover over $20,000$ languages, demonstrating that inheritance-based propagation substantially enhances \textsc{URIEL+}’s representation of low-resource languages.

\subsection{Lineage Imputation Lowers Sparsity}
\begin{figure}
    \centering
    \includegraphics[width=1.0\linewidth]{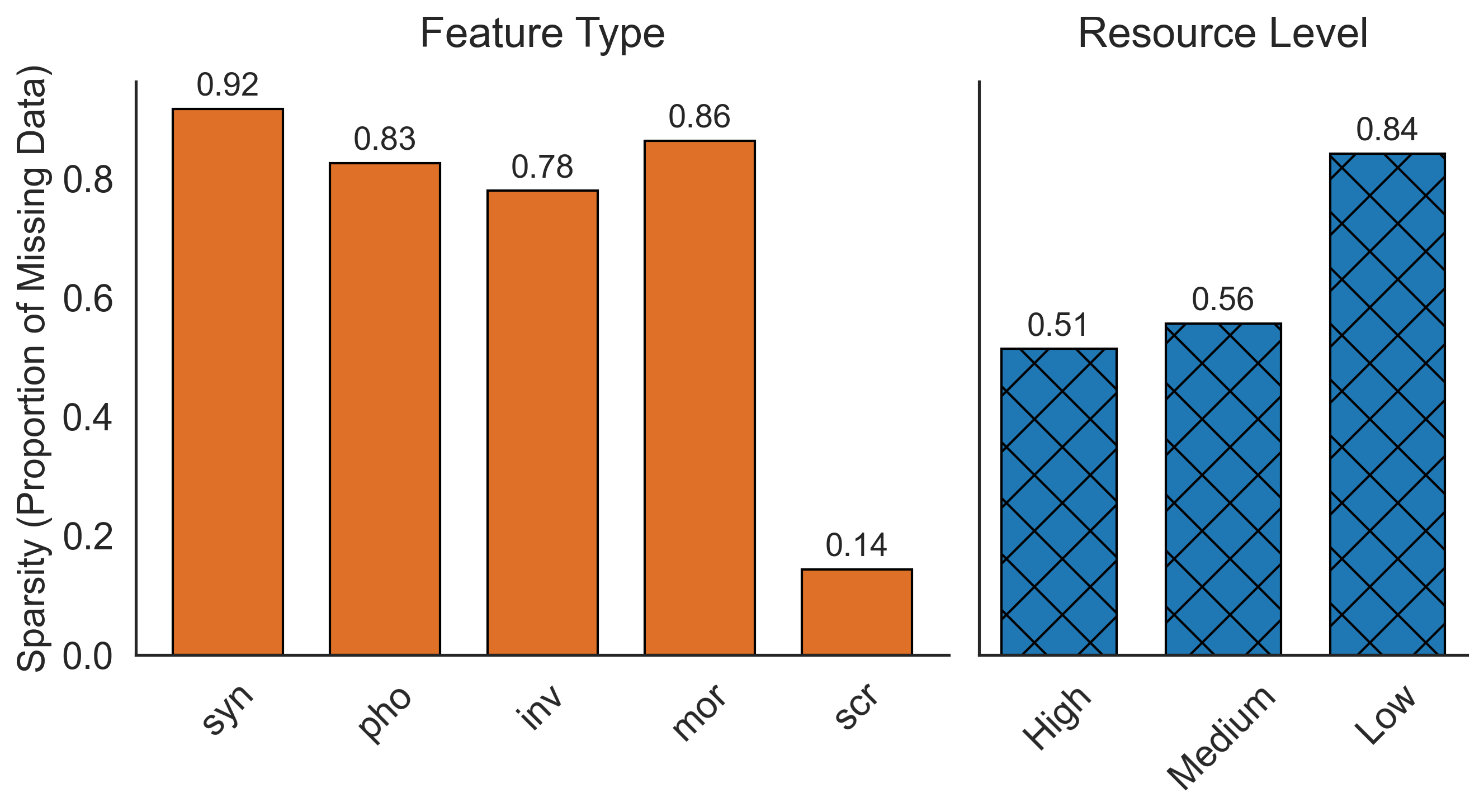}
    \caption{Sparsity of language vectors by type (syntax, phonological, inventory, morphological, script) and by resource level, \textbf{before} expanded lineage imputation.}
    \label{fig:sparsity-before}
\end{figure}

\begin{figure}
    \centering
    \includegraphics[width=1.0\linewidth]{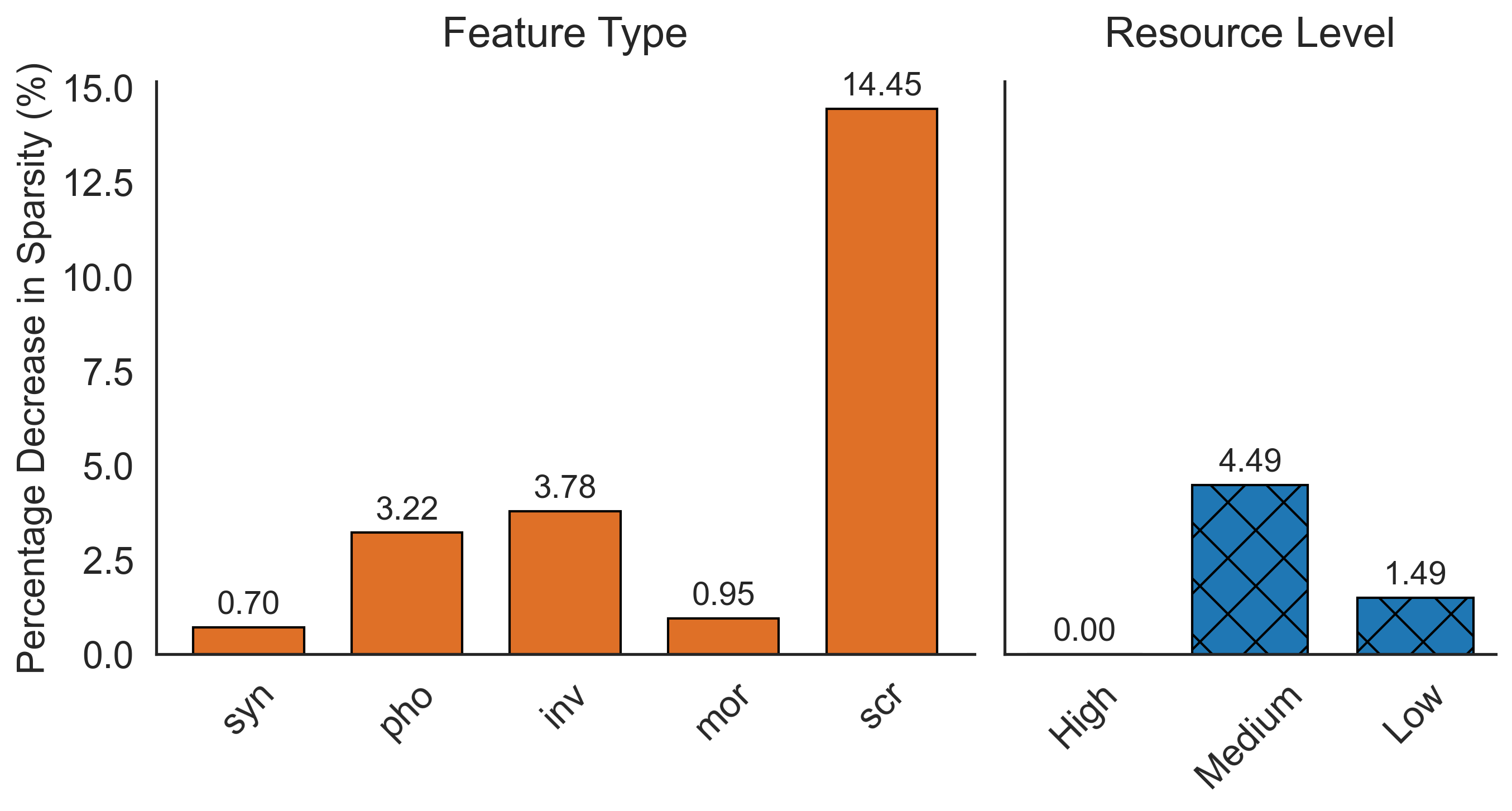}
    \caption{Percentage decrease in sparsity in language vectors by type (syntax, phonological, inventory, morphological, script) and by resource level, \textbf{after} expanded lineage imputation.}
    \label{fig:sparsity-after}
\end{figure}

A significant limiting factor to the reliability of \textsc{URIEL+} language vectors is their sparsity (i.e. the proportion of missing values). Breaking down typological vectors (into syntax, morphological, phonological and inventory vectors), Figure \ref{fig:sparsity-before} demonstrates the sparsity in language vectors by type, before expanded lineage imputation. While our findings confirm that the high sparsity in typological vectors is consistent in \textsc{URIEL+} across types ($78\% - 92\%$), script vectors notably possess a sparsity of only $14\%$. Coupled with their strong feature coverage, their low sparsity highlights the comprehensive nature of the data provided by script vectors. Furthermore, comparing the overall sparsity in typological and script vectors across the resource levels of languages, we find that data sparsity is skewed, as sparsity in LRLs ($84\%$) is substantially higher than in MRLs ($56\%$) and HRLs ($51\%$). The relatively high sparsity even among HRLs is mainly due to the large number of features included in \textsc{URIEL+}. \textsc{URIEL+} has $800$ features, many of whose values are derived from only one of the fourteen underlying data sources. Consequently, numerous features are sparsely populated across languages. For example, several features (e.g. S\_REGULARIZED\_REFLEXIVES\_PARADIGM, S\_DEMONSTRATIVES\_FOR\_DEFINITE\_ARTICLES, and S\_LIKE\_AS\_A\_QUOTATIVE\_PARTICLE) originate from the eWAVE \citelanguageresource{ewave} database and have recorded values for only around $70$ languages out of the thousands represented in the knowledge base.

Our expanded lineage imputation strategy provides a reliable method for filling missing entries with linguistically grounded values in languages. Figure \ref{fig:sparsity-after} shows the subsequent reduction in sparsity. This manifests in a reduction in sparsity (see Figure \ref{fig:sparsity-after}), particularly with sparsity in script vectors decreasing by $14\%$, i.e. to a sparsity of only $12\%$. However, comparing reductions in sparsity by vector type and resource level, expanded lineage imputation yields greater sparsity reductions where vectors are less sparse. In particular, sparsity reductions are minimal in syntax vectors and for LRLs, where sparsity was highest originally. This highlights how its efficacy is tied to the original sparsity of vectors since expanded lineage imputation relies on knowing feature values in parent languages.

\subsection{Scripts Provide Unique Information}
To justify the inclusion of script data, script distances should provide a different signal than typological, genetic, and geographic distances. In particular, they should not be substantially correlated with any of the other distance measures within \textsc{URIEL+}. Otherwise, script data would be redundant for the purpose of distance calculations--a primary use case of \textsc{URIEL+}.

To consider how similar script distances are to other distances in \textsc{URIEL+}, we study the correlation of a matrix containing all script distances with all other distance matrices in \textsc{URIEL+}. A distance matrix is a symmetric matrix, such that the $ij$th entry of the matrix is the distance between languages $i$ and $j$ in \textsc{URIEL+}. One can compute the Spearman correlation between the script matrix and any of the other distance matrices \cite{spearman1904}. To test the significance of the coefficient, a two-sided Mantel test \footnote{That is, a permutation test that builds a distribution of correlation statistics based on computing the Spearman correlation coefficient on permuted distance matrices \cite{mantel1967}.} is conducted.

 Mantel tests assume the lack of any kind of structured dependence (e.g. phylogenetic relationships between languages) between the two matrices \cite{guillot2013dismantling}. This is clearly not the case for language distances. Languages within the same language family are expected to have smaller distances than those with languages outside their family (see \citet{dunn2011nature}). To mitigate the violation of this assumption, we perform block permutation ($999$ many), where each permutation block \cite{winkler2015mlbp} is the top-level language family for a language, as reported in Glottolog. This alters the permutation procedure to allow us to account for phylogenetic dependencies between language distances. In total, there are $592$ language family blocks ($361$ of which are singletons) and approximately $95\%$ of languages fall into non-trivial language families (which mitigates discreteness caused by singletons). These tests are run on union-aggregated language vectors. For imputation, we run the correlation analysis with and without expanded lineage imputation, before using SoftImpute. We do not integrate any additional languages from Glottolog (i.e. beyond the $8,171$ already in \textsc{URIEL+}) because they do not have script data. The significance threshold we set is $\alpha = 0.05/7 \approx 0.007$. We apply a Bonferroni correction to account for conducting multiple hypothesis tests \citep{dunnMultipleComparisonsMeans1961}.

\begin{table}[!htp]
\centering
\small
\begin{tabular}{lrr}
\toprule
Distance & $\rho$ (No Lineage) & $\rho$ (Lineage) \\
\midrule
gen & $0.119$ & $0.114$ \\
inv & $0.059$ & $0.057$ \\
mor & $-0.026$ & $-0.035$\\
pho & $\mathbf{-0.041}$ & $-0.046$\\
syn & $\mathbf{0.060}$ & $0.051$\\
typ & $\mathbf{0.041}$ & $0.023$\\
geo & $\mathbf{-0.095}$ & $\mathbf{-0.095}$\\
\bottomrule
\end{tabular}
\caption{Spearman correlation ($\rho$) between language and script distance measures. ``Lineage'' denotes using expanded lineage imputation prior to SoftImpute. Significant results are \textbf{bolded}.}
\label{tab:lang-script-corr}
\end{table}

Table~\ref{tab:lang-script-corr} demonstrates that the magnitude of the correlation between script distances and all other distances in \textsc{URIEL+} is at most $0.119$, which is very minimal. This is irrespective of whether expanded lineage imputation was used or not. However, as seen in Table \ref{tab:lang-script-corr}, statistical significance is more common without expanded lineage imputation. This suggests that lineage imputation essentially acts as a phylogenetic regulariser that gives SoftImpute fewer degrees of freedom to infer patterns that might induce weak correlation with script distances. 

Even if some correlations are significant, all correlations are minimal. Taking geographic (post expanded lineage imputation) as an example, $\rho \approx -0.095$ is marginal. This means that the coefficient of variation, $R^2 \approx 0.009$, meaning that geography distances explain less than $1\%$ of the statistical variation in script distances. This supports our claim that script distances provide information that is largely \emph{orthogonal} to existing distance information within \textsc{URIEL+}, and hence, provide unique information.

\section{Evaluating our Additions}
We validated our \textsc{URIEL+} additions through two complementary evaluations: (a) testing imputation quality using the SoftImpute and lineage imputation algorithms and (b) applying language distances in cross-lingual transfer tasks using \textsc{LangRank}. 

\begin{table*}[!htp]
    \centering
    \resizebox{\textwidth}{!}{
    \begin{tabular}{llcccccc}
        \toprule
        \multirow{2}{*}{\textbf{Stage}} & \multirow{2}{*}{\textbf{Vector}} & \multicolumn{4}{c}{\textbf{Union-Agg}} & \multicolumn{2}{c}{\textbf{Average-Agg}} \\
        \cmidrule(lr){3-6} \cmidrule(lr){7-8}
         & & \textbf{Accuracy} & \textbf{Precision} & \textbf{Recall} & \textbf{F1} & \textbf{RMSE} & \textbf{MAE} \\
        \midrule
        \multirow{6}{*}{\makecell{$-$ Lineage}} 
            & typ & $0.8821$ & $0.8767$ & $0.7092$ & $0.7841$ & $0.2909$ & $0.1914$ \\
            & syn & $0.8410$ & $0.8496$ & $0.6650$ & $0.7461$ & $0.3357$ & $0.2467$ \\
            & pho & $0.9140$ & $0.9454$ & $0.8472$ & $0.8936$ & $0.2570$ & $0.1678$ \\
            & inv & $0.9432$ & $0.9267$ & $0.8245$ & $0.8726$ & $0.2015$ & $0.1040$ \\
            & mor & $0.8451$ & $0.8295$ & $0.5998$ & $0.6962$ & $0.3374$ & $0.2473$ \\
            & scr & $0.9939$ & $0.9949$ & $0.9826$ & $0.9887$ & $0.0769$ & $0.0248$ \\
        \midrule
        \multirow{6}{*}{\makecell{$+$ Lineage}} 
            & typ & $0.8880$ (\textcolor{darkblue}{$\uparrow 0.67\%$}) & $0.8821$ (\textcolor{darkblue}{$\uparrow 0.62\%$}) & $0.7238$ (\textcolor{darkblue}{$\uparrow 2.06\%$}) & $0.7951$ (\textcolor{darkblue}{$\uparrow 1.40\%$}) & $0.2845$ (\textcolor{darkblue}{$\downarrow 2.20\%$}) & $0.1876$ (\textcolor{darkblue}{$\downarrow 1.99\%$}) \\
            & syn & $0.8511$ (\textcolor{darkblue}{$\uparrow 1.20\%$}) & $0.8660$ (\textcolor{darkblue}{$\uparrow 1.93\%$}) & $0.6781$ (\textcolor{darkblue}{$\uparrow 1.97\%$}) & $0.7606$ (\textcolor{darkblue}{$\uparrow 1.94 \%$}) & $0.3262$ (\textcolor{darkblue}{$\downarrow 2.83\%$}) & $0.2399$ (\textcolor{darkblue}{$\downarrow 2.76\%$}) \\
            & pho & $0.9048$ (\textcolor{darkred}{$\downarrow 1.01\%$}) & $0.9180$ (\textcolor{darkred}{$\downarrow 2.90\%$}) & $0.8540$ (\textcolor{darkblue}{$\uparrow 0.80\%$}) & $0.8848$ (\textcolor{darkred}{$\downarrow 0.98 \%$}) & $0.2626$ (\textcolor{darkred}{$\uparrow 2.18\%$}) & $0.1691$ (\textcolor{darkred}{$\uparrow 0.77\%$}) \\
            & inv & $0.9436$ (\textcolor{darkblue}{$\uparrow 0.04\%$}) & $0.9242$ (\textcolor{darkred}{$\downarrow 0.27\%$}) & $0.8294$ (\textcolor{darkblue}{$\uparrow 0.59\%$}) & $0.8742$ (\textcolor{darkblue}{$\uparrow 0.18\%$}) & $0.2034$ (\textcolor{darkred}{$\uparrow 0.94\%$}) & $0.1074$ (\textcolor{darkred}{$\uparrow 3.27\%$}) \\
            & mor & $0.8545$ (\textcolor{darkblue}{$\uparrow 1.11\%$}) & $0.8319$ (\textcolor{darkblue}{$\uparrow 0.29\%$}) & $0.6312$ (\textcolor{darkblue}{$\uparrow 5.24\%$}) & $0.7178$ (\textcolor{darkblue}{$\uparrow 3.10\%$}) & $0.3260$ (\textcolor{darkblue}{$\downarrow 3.38\%$}) & $0.2366$ (\textcolor{darkblue}{$\downarrow 4.33\%$}) \\
            & scr & $\mathbf{0.9946}$ (\textcolor{darkblue}{$\uparrow 0.07\%$}) & $\mathbf{0.9957}$ (\textcolor{darkblue}{$\uparrow 0.08 \%$}) & $\mathbf{0.9842}$ (\textcolor{darkblue}{$\uparrow 0.16 \%$}) & $\mathbf{0.9899}$ (\textcolor{darkblue}{$\uparrow 0.12 \%$}) & $\mathbf{0.0740}$ (\textcolor{darkblue}{$\downarrow 3.77 \%$}) & $\mathbf{0.0242} $ (\textcolor{darkblue}{$\downarrow 2.42 \%$}) \\
        \bottomrule
    \end{tabular}
    }
    \caption{Performance of SoftImpute on typological and script vectors before ($-$) and after ($+$) expanded lineage imputation. Metrics are reported for union-aggregated and average-aggregated evaluation schemes. Values that are imputed through lineage imputation are excluded from being held out during imputation evaluation. \textbf{Bolded} values indicate the best score for each metric.}
    \label{tab:imputation-before-after-genetic-without-eval-results}
\end{table*}

\begin{table*}[!htp]
    \centering
    \small
    \setlength{\tabcolsep}{6pt}
    \renewcommand{\arraystretch}{1.1}
    \begin{tabular*}{\textwidth}{@{\extracolsep{\fill}}lcccccc@{}}
        \toprule
        \multirow{2}{*}{\textbf{Vector}} & \multicolumn{4}{c}{\textbf{Union-Agg}} & \multicolumn{2}{c}{\textbf{Average-Agg}} \\
        \cmidrule(lr){2-5} \cmidrule(lr){6-7}
        & \textbf{Accuracy} & \textbf{Precision} & \textbf{Recall} & \textbf{F1} & \textbf{RMSE} & \textbf{MAE} \\
        \midrule
        typ & $0.9886$ & $0.9823$ & $0.9757$ & $0.9790$ & $0.1068$ & $0.0114$ \\
        syn & $0.9925$ & $0.9902$ & $0.9892$ & $0.9897$ & $0.0863$ & $0.0075$ \\
        pho & $0.9879$ & $0.9863$ & $0.9846$ & $0.9854$ & $0.1099$ & $0.0121$ \\
        inv & $0.9870$ & $0.9768$ & $0.9634$ & $0.9701$ & $0.1139$ & $0.0130$ \\
        mor & $0.9937$ & $0.9856$ & $0.9948$ & $0.9902$ & $0.0793$ & $0.0063$ \\
        scr & \textbf{0.9987} & \textbf{0.9976} & \textbf{0.9974} & \textbf{0.9975} & \textbf{0.0366} & \textbf{0.0013} \\
        \bottomrule
    \end{tabular*}
    \caption{Performance of lineage imputation on typological and script vectors. Metrics are reported for union-aggregated and average-aggregated evaluation schemes. \textbf{Bolded} values indicate the best score for each metric.}
    \label{tab:imputation-before-after-lineage-results}
\end{table*}

\subsection{Imputation Quality Test}

\paragraph{Experimental Setup}
To assess the contribution of script vectors and expanded lineage imputation to reducing sparsity, we conducted two imputation quality tests following the methodology used in \textsc{URIEL+} \citelanguageresource{khan-etal-2025-uriel} and \citet{li2024comparison}. The procedure removes $20\%$ of known (i.e. non-missing) data, imputes these values, and compares predictions to the ground truth using F1 for binary features and root mean square error (i.e. RMSE) for continuous features. For our first test, we evaluated the SoftImpute algorithm. SoftImpute previously achieved the strongest performance of the three \textsc{URIEL+} imputation algorithms and was chosen for downstream and distance-alignment experiments \citelanguageresource{khan-etal-2025-uriel}. For this test, languages were pre-filled via our expanded lineage imputation procedure, however, filled values are not held out during imputation evaluation. This choice is to solely identify the quality of SoftImpute. With that said, we also report the results when we allow lineage-imputed values to be held out during imputation evaluation (i.e. treated as known), and the results of this can be found in Table~\ref{tab:imputation-before-after-genetic-with-eval-results} in Appendix \ref{app:imp-quantitative}.

For our second test, we evaluated solely the lineage imputation, skipping SoftImpute. Imputation was performed on aggregated representations (union aggregation, meaning feature values were merged using a logical OR operation; or average, meaning feature values were averaged).

\begin{table*}[!htp]
\centering
\setlength{\tabcolsep}{3pt}
\renewcommand{\arraystretch}{1.1}
\small
\begin{tabular}{lccccccccc}
\toprule
Setup & POS & DEP & XNLI & Taxi1500 & SIB200 & POS2 & DEP2 & MT & EL \\
\midrule
Baseline       & 18.5 & 73.0 & 67.6 & 23.3 & 8.38 & 33.8 & 36.3 & 35.9 & 72.1 \\
1             & \textbf{25.55} \up{38} & 71.6 \down{1.9} & 68.0 \up{0.6} & \textbf{25.51} \up{9.7} & 8.84 \up{5.5} & 34.1 \up{0.8} & 38.7 \up{6.6} & 31.7 \down{11.7} & 75.4 \up{4.5} \\
2             & 19.0 \up{2.8} & 73.0 \down{0.1} & 70.0 \up{3.6} & \textbf{26.1} \up{12} & 7.58 \down{9.5} & 32.1 \down{5} & 36.1 \down{0.5} & 33.1 \down{7.6} & 69.9 \down{3.1} \\
1 + 2         & \textbf{27.03} \up{46} & 72.3 \down{1} & 67.2 \down{0.5} & \textbf{26.92} \up{15.7} & 8.40 \up{0.2} & 33.2 \down{1.7} & 36.5 \up{0.6} & 32.0 \down{10.8} & 66.9 \down{7.2} \\
3             & 17.0 \down{8} & 73.0 & 67.9 \up{0.4} & 22.7 \down{2.4} & 9.76 \up{16.5} & 32.5 \down{3.7} & 38.3 \up{5.5} & 35.9 \up{0.1} & 67.8 \down{5.9} \\
1 + 3         & 21.3 \up{14.8} & 73.4 \up{0.5} & 67.5 \down{0.1} & \textbf{25.33} \up{8.9} & 9.01 \up{7.5} & 34.0 \up{0.7} & 37.6 \up{3.6} & 33.4 \down{7} & 69.7 \down{3.3} \\
2 + 3         & 13.1 \down{29.1} & 74.5 \up{2.1} & 71.5 \up{5.7} & 23.5 \up{1.1} & 9.87 \up{17.8} & 36.3 \up{7.3} & 39.1 \up{7.9} & 34.9 \down{2.8} & 61.5 \down{14.7} \\
1 + 2 + 3     & 18.6 \up{0.5} & 72.6 \down{0.5} & 70.9 \up{4.9} & 25.0 \up{7.5} & 9.79 \up{16.8} & 36.3 \up{7.5} & 38.0 \up{4.9} & \textbf{30.2} \down{15.8} & 62.8 \down{12.9} \\
\bottomrule
\end{tabular}
\caption{NDCG@3 scores in \textsc{LangRank} tasks under our ablation study. Percentage changes are relative to the baseline. \textbf{Bolded} values indicate $p<0.05$ in comparison to the baseline. Baseline is SoftImpute. 1 = Script; 2 = Glottolog integration; 3 = Lineage Impute.}
\label{tab:task_results}
\end{table*}

\paragraph{Results}
Table~\ref{tab:imputation-before-after-genetic-without-eval-results} shows that SoftImpute achieves near-perfect performance on script vectors, with accuracy and F1 above $0.98$ across all configurations. These results substantially exceed those for all types of typological vectors, indicating that script features are both more complete and predictable within genealogical lineages.

There are two factors that explain this result. First, script data are inherently less sparse (i.e. before imputation, script vectors cover $7,488$ languages (nearly twice the most complete typological distance type, which is syntactic), and after imputation, coverage exceeds $20,000$ languages). Secondly, writing system properties (e.g. the complexity of characters) are highly stable within language families \cite{mitonGraphicComplexityWriting2021}, providing strong structural cues for imputation models.

Applying expanded lineage imputation improves most metrics across every vector type, with the largest gains for typological features (in particular morphological) under union aggregation. For instance, recall increases by \textcolor{black}{$2.06\%$} overall (\textcolor{black}{$5.24\%$} for morphological) and F1 increases by \textcolor{black}{$1.40\%$} (\textcolor{black}{$3.10\%$} for morphological), gains attributable to their initial sparsity. These results confirm that expanded lineage imputation improves imputation quality and is effective in reducing sparsity. 

The exceptions are phonological and inventory features. Phonological features decline in accuracy (\textcolor{darkred}{$\downarrow 1.01\%$}), precision (\textcolor{darkred}{$\downarrow 2.90\%$}), and F1 (\textcolor{darkred}{$\downarrow 0.98\%$}) following expanded lineage imputation, while inventory features decline in precision (\textcolor{darkred}{$\downarrow 0.27\%$}). Both feature types also decline under average aggregation: phonological features worsen by \textcolor{black}{$2.18\%$} in RMSE and \textcolor{black}{$0.77\%$} in MAE, and inventory features by \textcolor{black}{$0.94\%$} in RMSE and \textcolor{black}{$3.27\%$} in MAE. This suggests that lineage propagation may not be well-suited for these feature types. Phonological features encode how sounds are produced, including manner and place of articulation. Meanwhile, inventory features capture the phonetic sound inventories of languages. Neither set of properties may transfer consistently across related languages. Supporting this idea, Table~\ref{tab:imputation-before-after-lineage-results} shows that phonological and inventory features also show the weakest performance of all vector types when evaluating lineage imputation quality alone, achieving the lowest scores across all metrics under both union and average aggregation.

Table~\ref{tab:imputation-before-after-lineage-results} shows that lineage imputation performs strong across both typological and script vectors, with script vectors outperforming typological ones on every metric. This is likely because, as mentioned earlier, writing systems are more consistently inherited within language families, making parent-to-child propagation a particularly reliable assumption for script data. 

Despite trailing script vectors, typological performance remains exceptionally high (F1 = $0.9790$, Accuracy = $0.9886$). The higher RMSE under average aggregation ($0.1068$ compared to $0.0366$) suggests more continuous prediction error for typological features, which is expected since typological features have greater cross-linguistic variability.


These results confirm that our additions not only increase coverage but also improve the internal consistency and predictability of \textsc{URIEL+}, demonstrating the combined benefit of script vectors and expanded lineage imputation.

\subsection{Benchmarking Cross-Lingual Transfer}
\paragraph{Evaluation Setup} Given the common application of language distances in transfer language selection \cite{blaschke-etal-2025-analyzing, rice-etal-2025-untangling, ng-etal-2025-less, ng2026modality}, we evaluate the impact of our additions on cross-lingual transfer using the \textsc{LangRank} framework \cite{lin-etal-2019-choosing}. \textsc{LangRank}, which ranks potential transfer languages for multilingual NLP tasks using gradient-boosted decision trees trained on language distances. Following the framework of \citet{ng2026modality}\footnote{We employ the same \textsc{LangRank} hyperparameters and the same set of tasks.}, we apply \textsc{LangRank} to select transfer languages across nine representative sub-tasks: machine translation (MT), entity linking (EL), part-of-speech tagging (POS 1 \& 2), dependency parsing (DEP 1 \& 2), natural language inference (XNLI \cite{conneau-etal-2018-xnli}), and topic classification (Taxi1500 \cite{ma-etal-2025-taxi1500}, SIB200 \cite{adelani-etal-2024-sib}). This framework measures transfer language selection quality (i.e. how well language distances rank potential source languages for a given target language) and measures transfer language selection quality, specifically, how well language distances rank potential source languages for a given target language. The \textit{Normalized Discounted Cumulative Gain (NDCG@3)} metric evaluates whether our improved distance calculations help identify the top-3 most promising source languages for training, compared to the empirically determined optimal source. 

We conduct an ablation study to isolate the effect of each contribution and to quantify its practical impact on transfer predictions. Specifically, we train \textsc{LangRank} on \textsc{URIEL+} distances following SoftImpute, under three configurations: (1) incorporating script distances, (2) integrating languages from Glottolog, and (3) conducting expanded lineage imputation prior to SoftImpute. Transfer Performance is evaluated using the Normalised Discounted Cumulative Gain at rank 3 (NDCG@3), which measures alignment between \textsc{LangRank's} predicted transfer language rankings and the optimal language rankings.task transferability.

\paragraph{Results} Table \ref{tab:task_results} demonstrates \textsc{LangRank} performance across ablations. Overall, performance gains are task-specific, appearing incan be found in all but two tasks studied (MT, EL). In particular, incorporating script distances significantly improves \textsc{LangRank} transfer predictions in part-of-speech tagging and Taxi1500 topic classification. Simultaneously, expanded lineage imputation yields further improvements in SIB200 topic classification and DEP 2. Given the high language coverage in these tasks ($799$ in Taxi1500, $197$ in SIB200 \cite{ng2026modality}), most of which are low-resource, these improvements demonstrate how our extensions to low-resource language coverage support cross-lingual transfer.

Generally, while performance gains underscore the utility of our contributions in choosing transfer languages, improvements remain modest and fluctuate between tasks and setups. Simply incorporating script distance yields an average performance improvement of $6\%$, the highest among all other setups. Moreover, many improvements do not reach statistical significance, and our additions exhibit variable performance impacts across tasks and setups. However, our downstream evaluation has limited scope for assessing our additions. While we expand data coverage in URIEL+ for mostly low-resource languages, the downstream tasks primarily cover higher-resource languages already attested in URIEL+ previously. Thus, our downstream evaluation on transfer performance does not directly benchmark the expanded language coverage that we contribute. Therefore, we turn to task-level patterns to better characterize where our additions are effective.

The variation in performance gains across tasks is informative and it highlights the conditions under which our additions are most effective. For instance, script vectors significantly improve transfer language selection for POS tagging, where differences in writing system hamper transfer performance in morphological tasks \cite{murikinati-etal-2020-transliteration}, and where languages sharing a script tend to share tokenisation behaviour \cite{xhelili-etal-2024-breaking, zhuang-etal-2025-enhancing}. Simultaneously, e\textcolor{red}{}xpanded lineage imputation yields gains in tasks which include a larger proportion of medium- and low-resource languages (SIB200, POS2 and DEP2), consistent with the reduced data sparsity of languages in those resource levels. Overall, our additions tend to improve transfer where they provide information relevant to the task. This finding is consistent with \citet{ng2026modality}, and underscores the importance of multi-faceted approaches to improving data coverage in URIEL+.

\section{Conclusion}
Our paper introduces a systematic approach to reducing sparsity in the \textsc{URIEL+} linguistic knowledge base. By introducing script vectors for $7,488$ languages, incorporating $18,710$ languages from Glottolog, and extending lineage imputation to $26,449$ languages, we substantially improve data coverage, particularly for low-resource languages. Script vectors provide explicit representations of writing systems, supporting more reliable cross-lingual modelling. The Glottolog additions expand the knowledge base’s genealogical breadth, while expanded lineage imputation propagates typological and script features through these genealogies, filling gaps for low-resource languages. These additions reduce feature sparsity, increase language coverage, and improve imputation quality metrics. Furthermore, we benchmark the additions applicability in cross-lingual transfer. By releasing our improvements, we provide the NLP community with a more complete and equitable foundation for multilingual research, particularly benefiting under-represented languages. We have provided a new release of URIEL+, with our additions found here: \url{https://github.com/LeeLanguageLab/URIELPlus}.

\section{Future Work}
The lineage imputation scheme we explore is generally motivated by the agreement in feature values between parent and child languages. However, it is rather simple. We plan to explore more advanced imputation schemes that take advantage of the tree-like structure of language phylogeny. 

Furthermore, since \textsc{URIEL+} is not synchronized with Glottolog and ScriptSource, we intend to periodically update it using new releases from these sources to maintain alignment with current language and script metadata.

\section{Limitations}
Our work relies on existing data sources such as ScriptSource and Glottolog. Therefore, it is subject to any biases or inaccuracies present in them. We could not perfectly replicate the baseline imputation quality results from \textsc{URIEL+}, despite running the same code provided by \citelanguageresource{khan-etal-2025-uriel} and achieving the same feature coverage prior to imputation (see syntactic, phonological, inventory, and morphological feature coverage before expanded lineage imputation in Figure~\ref{fig:feature_coverage_resource_level}). Consequently, the distances used in our downstream experiments differed slightly, leading to minor variations in downstream performance. 

Furthermore, the features themselves have inherent limitations. Script similarity does not always equate to the deep linguistic similarity required for complex downstream tasks. For example, script vectors correctly capture that Persian and Arabic share script features. However, Persian is an Indo-European language while Arabic is Semitic. For a task like machine translation, over-reliance on the script signal could incorrectly suggest a stronger linguistic relationship than exists, potentially adding a misleading signal.

Similarly, our expanded lineage imputation operates on the heuristic that dialects share features with their parent languages. While broadly effective, this assumption can be flawed. A dialect may have evolved a unique feature or, through language contact, lost a feature that its parent retained. In such cases, the imputation would be incorrect, adding a faulty data point that could negatively impact downstream tasks. Furthermore, any errors or biases in the documented parent language features are systematically inherited by all descendants. This risks potentially amplifying inaccuracies in our dataset, particularly for parent language where prestige varieties are particularly favoured, as is common for many of the low-resource families our method targets \cite{joshi-etal-2020-state}.

Finally, adding $18,710$ languages from Glottolog to \textsc{URIEL+}, each with genetic, geographic, typological (syntactic, phonological, inventory, and morphological), and script vectors, substantially increases the time required to integrate all linguistic data sources in \textsc{URIEL+} (e.g. from a few minutes to often around an hour). 

\section{Ethics Statement}
This work builds upon publicly available linguistic resources, including URIEL+, Glottolog, and ScriptSource. \textsc{URIEL+} and Glottolog are both publicly viewable and accessible, whereas ScriptSource is publicly viewable but not openly accessible for bulk data extraction or redistribution. Access to ScriptSource data for this research was obtained with the explicit permission and assistance of the resource's administrator, to ensure compliant and responsible use. Furthermore, we received the permission of the \textsc{URIEL+} authors to contribute and extend their work.

\begin{table*}[!htp]
    \centering
    \resizebox{\textwidth}{!}{
    \begin{tabular}{llcccccc}
        \toprule
        \multirow{2}{*}{\textbf{Stage}} & \multirow{2}{*}{\textbf{Vector}} & \multicolumn{4}{c}{\textbf{Union-Agg}} & \multicolumn{2}{c}{\textbf{Average-Agg}} \\
        \cmidrule(lr){3-6} \cmidrule(lr){7-8}
         & & \textbf{Accuracy} & \textbf{Precision} & \textbf{Recall} & \textbf{F1} & \textbf{RMSE} & \textbf{MAE} \\
        \midrule
        \multirow{6}{*}{\makecell{$-$ Lineage}} 
            & typ & $0.8821$ & $0.8767$ & $0.7092$ & $0.7841$ & $0.2909$ & $0.1914$ \\
            & syn & $0.8410$ & $0.8496$ & $0.6650$ & $0.7461$ & $0.3357$ & $0.2467$ \\
            & pho & $0.9140$ & $0.9454$ & $0.8472$ & $0.8936$ & $0.2570$ & $0.1678$ \\
            & inv & $0.9432$ & $0.9267$ & $0.8245$ & $0.8726$ & $0.2015$ & $0.1040$ \\
            & mor & $0.8451$ & $0.8295$ & $0.5998$ & $0.6962$ & $0.3374$ & $0.2473$ \\
            & scr & $0.9939$ & $0.9949$ & $0.9826$ & $0.9887$ & $0.0769$ & $0.0248$ \\
        \midrule
        \multirow{6}{*}{\makecell{$+$ Lineage}} 
            & typ & $0.9437$ (\textcolor{darkblue}{$\uparrow 6.98\%$}) & $0.9453$ (\textcolor{darkblue}{$\uparrow 7.82\%$}) & $0.8477$ (\textcolor{darkblue}{$\uparrow 19.53\%$}) & $0.8938$ (\textcolor{darkblue}{$\uparrow 13.99\%$}) & $0.2133$ (\textcolor{darkblue}{$\downarrow 26.68\%$}) & $0.1250$ (\textcolor{darkblue}{$\downarrow 34.69\%$}) \\
            & syn & $0.8927$ (\textcolor{darkblue}{$\uparrow 6.15\%$}) & $0.9134$ (\textcolor{darkblue}{$\uparrow 7.51\%$}) & $0.7716$ (\textcolor{darkblue}{$\uparrow 16.03\%$}) & $0.8366$ (\textcolor{darkblue}{$\uparrow 12.13\%$}) & $0.2873$ (\textcolor{darkblue}{$\downarrow 14.42\%$}) & $0.2058$ (\textcolor{darkblue}{$\downarrow 16.58\%$}) \\
            & pho & $0.9649$ (\textcolor{darkblue}{$\uparrow 5.57\%$}) & $0.9729$ (\textcolor{darkblue}{$\uparrow 2.91\%$}) & $0.9420$ (\textcolor{darkblue}{$\uparrow 11.19\%$}) & $0.9572$ (\textcolor{darkblue}{$\uparrow 7.12\%$}) & $0.1833$ (\textcolor{darkblue}{$\downarrow 28.68\%$}) & $0.1087$ (\textcolor{darkblue}{$\downarrow 35.22\%$}) \\
            & inv & $0.9715$ (\textcolor{darkblue}{$\uparrow 3.00\%$}) & $0.9668$ (\textcolor{darkblue}{$\uparrow 4.33\%$}) & $0.9018$ (\textcolor{darkblue}{$\uparrow 9.38\%$}) & $0.9332$ (\textcolor{darkblue}{$\uparrow 6.94\%$}) & $0.1567$ (\textcolor{darkblue}{$\downarrow 22.23\%$}) & $0.0786$ (\textcolor{darkblue}{$\downarrow 24.42\%$}) \\
            & mor & $0.8885$ (\textcolor{darkblue}{$\uparrow 5.14\%$}) & $0.8962$ (\textcolor{darkblue}{$\uparrow 8.04\%$}) & $0.7223$ (\textcolor{darkblue}{$\uparrow 20.42\%$}) & $0.7999$ (\textcolor{darkblue}{$\uparrow 14.90\%$}) & $0.2947$ (\textcolor{darkblue}{$\downarrow 12.66\%$}) & $0.2107$ (\textcolor{darkblue}{$\downarrow 14.80\%$}) \\
            & scr & $\mathbf{0.9949}$ (\textcolor{darkblue}{$\uparrow 0.10\%$}) & $\mathbf{0.9958}$ (\textcolor{darkblue}{$\uparrow 0.09\%$}) & $\mathbf{0.9851}$ (\textcolor{darkblue}{$\uparrow 0.25\%$}) & $\mathbf{0.9904}$ (\textcolor{darkblue}{$\uparrow 0.17\%$}) & $\mathbf{0.0729}$ (\textcolor{darkblue}{$\downarrow 5.20\%$}) & $\mathbf{0.0240}$ (\textcolor{darkblue}{$\downarrow 3.23\%$}) \\
        \bottomrule
    \end{tabular}
    }
    \caption{Performance of SoftImpute on typological and script vectors before ($-$) and after ($+$) expanded lineage imputation. Metrics are reported for union-aggregated and average-aggregated evaluation schemes. Values that are imputed through lineage imputation are included in the imputation evaluation. \textbf{Bolded} values indicate the best score for each metric.}
    \label{tab:imputation-before-after-genetic-with-eval-results}
\end{table*}

All data used in this work describe linguistic structures and writing systems, without involving personally identifiable or user-generated content. Our primary objective is to address data sparsity in multilingual linguistic knowledge bases by systematically enriching low-resource features rather than collecting new human data. While this enrichment improves coverage, automatic feature propagation through expanded lineage imputation may inadvertently reinforce existing typological biases from well-documented languages. To mitigate this, we ensured that the imputation procedure was transparent and reproducible and we release all derived resources and code under an open license to promote transparency and community oversight.

\section*{Acknowledgements}
We thank Martin Raymond and ScriptSource for providing us the script feature data, Khasir Hean for his feedback on the expanded lineage imputation, and Patrick Littell for suggesting the use of ScriptSource for constructing our script vectors.

\appendix

\section{Additional Results for the Imputation Quality Test}\label{app:imp-quantitative}

In addition to the imputation quality tests done in Tables~\ref{tab:imputation-before-after-genetic-without-eval-results} and~\ref{tab:imputation-before-after-lineage-results}, another imputation quality test was done in Table~\ref{tab:imputation-before-after-genetic-with-eval-results}. This test has values filled with the expanded lineage imputation procedure held out during imputation evaluation. The imputation results from this test are the results that users of \textsc{URIEL+} would see when they integrate all databases and impute with both lineage imputation and SoftImpute and are the best imputation results of all quality tests done.

Table~\ref{tab:imputation-before-after-genetic-with-eval-results} shows that lineage imputation consistently improves SoftImpute performance across all vector types and aggregations. The gains are most pronounced for typological vectors, with recall increasing from $0.7092$ to $0.8477$ (\textcolor{darkblue}{$\uparrow 19.53\%$}), F1 increasing from $0.7841$ to $0.8938$ (\textcolor{darkblue}{$\uparrow 13.99\%$}), RMSE dropping from $0.2964$ to $0.2175$ (\textcolor{darkblue}{$\downarrow 26.68\%$}), and MAE dropping from $0.1961$ to $0.1282$ (\textcolor{darkblue}{$\downarrow 34.69\%$}), suggesting that lineage imputation meaningfully fills in feature gaps that SoftImpute alone struggles with. We can see even stronger individual improvements with recall and F1 for morphological vectors increasing from $0.5998$ to $0.7223$ (\textcolor{darkblue}{$\uparrow 20.42\%$}) and $0.6962$ to $0.7999$ (\textcolor{darkblue}{$\uparrow 14.90\%$}), respectively. We see RMSE and MAE for phonological vectors decreasing from $0.2570$ to $0.1833$ (\textcolor{darkblue}{$\downarrow 28.68\%$}) and $0.1678$ to $0.1087$ (\textcolor{darkblue}{$\downarrow 35.22\%$}), respectively. Script vectors show more modest but consistent improvements across all metrics, which is unsurprising given their already strong baseline performance.

The better performance when lineage imputation is considered alongside SoftImpute is because lineage imputation fills values that are highly predictable from family structure--cases where a child language inherits a feature directly from its parent. These are structurally easy cases, and SoftImpute tends to predict them well precisely because related languages cluster together in feature space. Including them in the evaluation therefore inflates scores relative to the harder, non-lineage-covered positions, which explains why the typological gains are so large. Typological features have broader lineage coverage, meaning more of these easy, family-consistent positions are added back into the evaluation pool.




\section{Bibliographical References}\label{sec:reference}

\bibliographystyle{lrec2026-natbib}
\bibliography{main}

\section{Language Resource References}

\label{lr:ref}
\bibliographystylelanguageresource{lrec2026-natbib}
\bibliographylanguageresource{languageresource}

\end{document}